\documentclass[a4paper]{article}

\usepackage{INTERSPEECH2020}
\usepackage{booktabs}
\usepackage{enumitem}
\usepackage{dsfont}
\usepackage{algorithm}
\usepackage{algpseudocode} 
\usepackage{amsmath}
\usepackage{subcaption}
\usepackage[symbol]{footmisc}

\DeclareMathOperator*{\argmax}{argmax}

\title{Prototypical Q Networks for Automatic Conversational Diagnosis\\ and Few-Shot New Disease Adaption}
\name{Hongyin Luo$^1$, Shang-Wen Li$^2$\thanks{Work done before the second author joined Amazon}, James Glass$^1$}
\address{
  $^1$MIT CSAIL\\
  $^2$Amazon AI}
\email{hyluo@mit.edu, shangwel@amazon.com, glass@mit.edu}

\begin{document}

\maketitle
\begin{abstract}
Spoken dialog systems have seen applications in many domains, including medical for automatic conversational diagnosis. State-of-the-art dialog managers are usually driven by deep reinforcement learning models, such as deep Q networks (DQNs), which learn by interacting with a simulator to explore the entire action space since real conversations are limited. However, the DQN-based automatic diagnosis models do not achieve satisfying performances when adapted to new, unseen diseases with only a few training samples. In this work, we propose the Prototypical Q Networks (ProtoQN) as the dialog manager for the automatic diagnosis systems. The model calculates prototype embeddings with real conversations between doctors and patients, learning from them and simulator-augmented dialogs more efficiently. We create both supervised and few-shot learning tasks with the Muzhi corpus. Experiments showed that the ProtoQN significantly outperformed the baseline DQN model in both supervised and few-shot learning scenarios, and achieves state-of-the-art few-shot learning performances. 
% Deep reinforcement learning models, for example, deep Q networks (DQNs) have been applied for automatic conversational diagnosis. The models learn dialog policies by interacting with a user simulator, which stores symptoms and ground-truth disease labels. However, the models cannot learn from real doctor-patient conversations, and the DQNs do not achieve satisfying performance when adapted to new, unseen diseases with only a few training samples, especially when the disease is hard to be diagnosed. In this work, we propose the Prototypical Q Networks as the dialog manager for the automatic diagnosis systems. The model calculates prototypes with real dialogs between doctors and patients, making better use of the training data. We create both supervised learning and few-shot learning tasks with the Muzhi corpus. Experiments showed that the Prototypical Q Networks significantly outperformed the baseline DQN models in both supervised and few-shot learning scenarios, and achieves state-of-the-art few-shot learning performances.
\end{abstract}


\section{Introduction}
\begin{figure}[h]
\centering
\includegraphics[width=0.4\textwidth]{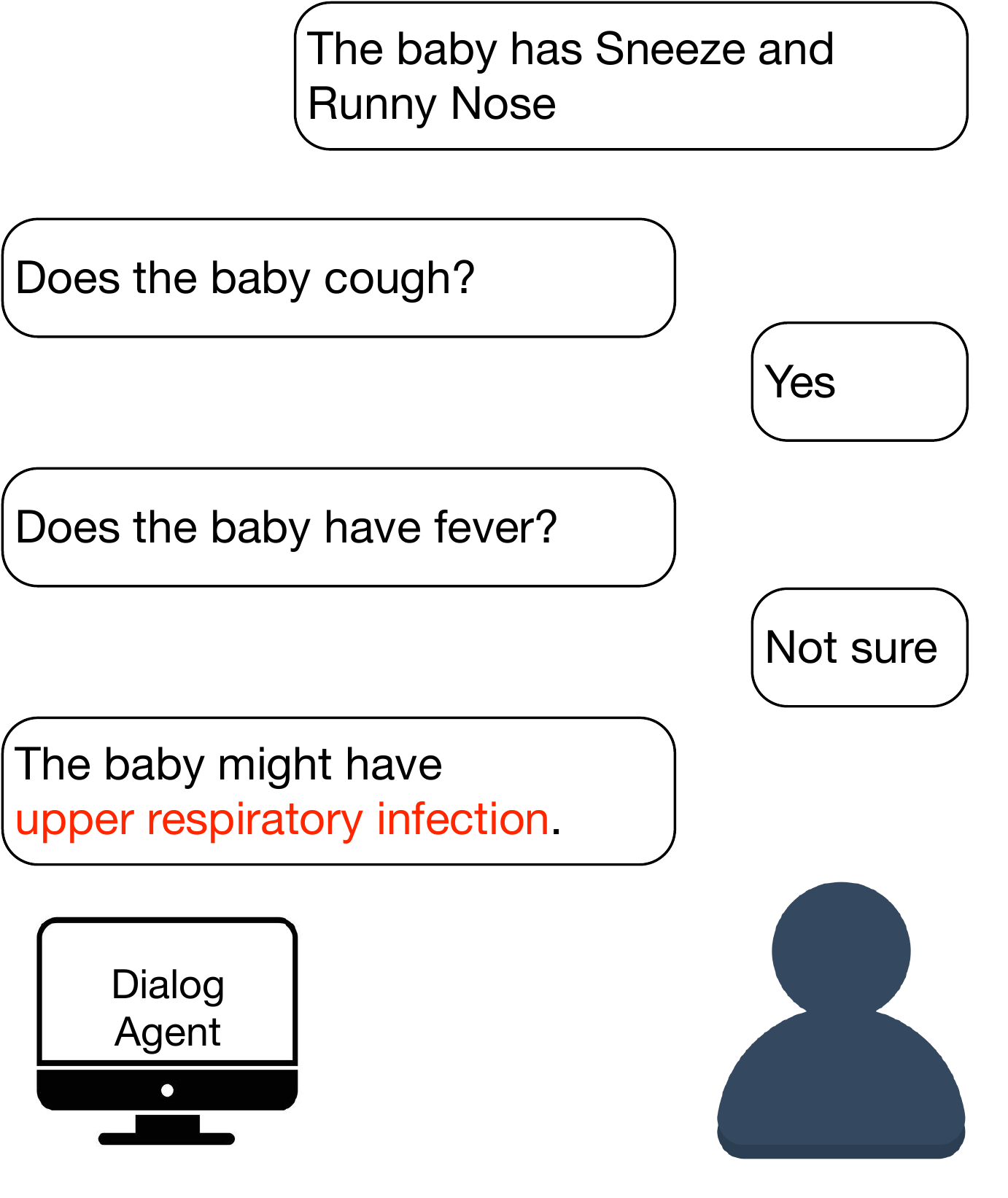}
\caption{An example of a dialog between a doctor and a user. Firstly, the user provides a self report. Then the agent conduct a dialog by requesting symptoms and conclude by making a decision about the disease.}
\label{fig:ad}
\end{figure}

Recently spoken dialog systems have been a popular research topic in human language technology (HLT) area with various applications. Among these applications, dialog systems for clinical conversation (i.e., medical bot) is a rising direction for its widespread and impactful use \cite{wei2018task}. Medical bot assists medical practitioners to converse with patients, collect information about their symptoms, physical and mental conditions, or even make suggestions on diagnosis. The bot has significant potential to make the diagnostic procedure more efficient. An example for automatic diagnosis dialog system is shown in Figure \ref{fig:ad}. Starting from a self-report, the medical bot collects and distills symptom information before it makes the disease prediction.
% While clinical diagnosis are made by the doctors, according to their medical knowledge and symptom information of patients, collecting the information, sometimes via medical examinations, heavily relies on self-report of patients (e.g., ``My stomach does not feel good'' or ``My kid has a fever'') and the follow-up dialog between doctor and patients. With the self-report and follow-up dialog, doctors distill information about the patient and make clinical decisions. A medical bot for automatic conversational diagnosis systems simulate such process and predict the disease based on collected information \cite{wei2018task}. An example of such dialog is shown in Figure \ref{fig:ad}.
%
%
% \begin{table}[]
% \caption{An example of a dialog between a doctor and a user. Firstly, the user provides a self report. Then the agent conduct a dialog by requesting symptoms and conclude by making a decision about the disease.}
% \label{tab:ad}
% \centering
% \begin{tabular}{@{}ll@{}}
% \toprule
% \multicolumn{2}{c}{\textbf{1. Self Report}}                        \\
% \multicolumn{2}{l}{The baby has Sneeze and Runing Nose}         \\ \midrule
% \multicolumn{2}{c}{\textbf{2. Dialog}}                             \\
% Agent: Does baby cough?               & User: Yes               \\
% Agent: Does baby have fever?          & User: Not sure          \\ \midrule
% \multicolumn{2}{c}{\textbf{3. Diagnosis}}                          \\
% \multicolumn{2}{l}{Baby might have upper respiratory infection} \\ \bottomrule
% \end{tabular}
% \end{table}

One of the core challenges of building such a dialog system is design and train a dialog policy manager that can reason and decide the action to take based on the understanding of user intentions and conversation context. It is more challenging for medical bot because of the need of integrating medical knowledge for reasoning and decision making. \cite{wei2018task} proposed a reinforcement learning (RL) framework with a multi-layer perceptron deep Q networks (MLP-DQN) \cite{mnih2013playing}. \cite{xu2019end} extended the study by enhanceing the DQN with hand-crafted features among diseases and symptoms, generated from the training set. However, both models cannot directly learn from real doctor-patient  conversations.  For RL agent to fully explore the entire action space, the agent can only learn by interacting with rule-based user simulators, which can not learn from the dialog histories between real doctors and patients.

Another difficulty faced by RL-based manager is adapting trained policy to new tasks (e.g., adapt trained medical bot to serve new diseases), since adaptation data is usually limited and hard to collect. On the other hand, Meta-learning algorithms are proposed to improve the model performance when training or adaptation data is limited. \cite{finn2017model,snell2017prototypical}. Since both many- and few-shot learning heavily depend on the quality of learned representations, these studies encouraged us to combine meta-learning and deep reinforcement learning models to improve the dialog agents for automatic diagnosis in both scenarios by learning better representations of dialog actions and domain knowledge. 

In this work, we propose prototypical Q networks (ProtoQN) borrowing the ideas of prototypical networks \cite{snell2017prototypical} and matching networks \cite{vinyals2016matching}. We evaluate the model in the medical dialog domain, since it is important and medical conversation usually suffers from data scarcity. The model makes fully use of real doctor-patient conversations by calculating prototype disease and symptom embeddings by encoding the dialog histories in the training set. Experiment results have shown that by learning a shared prototype embedding space, the ProtoQN outperforms MLP-DQN under both experiment settings. The experiments in this paper will focus on medial dialog to show benefit of proposed method, but we believe the conclusion can be generalized to other domains, since we do not use handcrafted features or external domain specific information.

\section{Related Work}

\subsection{Deep Q Networks}
The deep Q networks (DQNs) are proposed in \cite{mnih2013playing} for handling Atari video games. \cite{silver2016mastering} proposed a deep reinforcement learning architecture for mastering the game of GO. In the area of dialog systems, the DQN is a popular model for building dialog managers and learning dialog policies \cite{zhao2016towards,yang2017end,lipton2016efficient,peng2017composite}.

DQN is usually implemented as the following. The goal of the network is estimating $ q(s, a) $, the Q value of taking action $a$ at state $s$. At each time step, an DQN-based agent selects an action with a $\epsilon$-greedy strategy, i.e., epsilon \% of the time selecting action with largest Q value given current state, and picking a random action for the rest. Meanwhile, the transition of the current step, $(S_t, A_t, R_t, \gamma, S_{t+1})$, is added to a memory buffer for future learning \cite{lin1992self}. Here, $S_t$ is the state at time $t$, $A_t$ is the action taken at time $t$, $R_{t+1}$ stands for the immediate reward at time step $t+1$, and $\gamma$ is a discount rate. The objective function for training the neural network is
\[
L = (R_t + \gamma max_{a'} q_{\bar{\theta}} (S_{t+1}, a') - q_{\theta}(S_t, A_t))^2
\]
where $\theta$ stands for the parameters set of the current network and $\bar{\theta}$ is the parameters of the target network. The parameters are updated by stochastic gradient descent (SGD).

\subsection{Spoken Dialog Systems}
Task-oriented spoken dialog systems aim at completing a specific task by interacting with users through natural language. Conventionally a dialog manager is built to learn the dialog policy, which decides actions by reasoning from dialog state (the combined representation of user intentions and context) \cite{papineni2001natural,scheffler2002automatic,young2010hidden}. Dialog management is often formulated as a partially observable Markov decision process (POMDP), and solved as a reinforcement learning (RL) problem \cite{young2013pomdp}. As of late, many state-of-the-art dialog systems achieve satisfactory results by leveraging DQNs \cite{mnih2013playing,silver2014deterministic}, to learn policy and manage dialogs \cite{zhao2016towards}.

Typical applications include flight booking \cite{seneff2000dialogue}, movie recommendation \cite{dodge2015evaluating}, restaurant reservation \cite{bordes2016learning}, and vision grounding \cite{chattopadhyay2017evaluating}. Medical bot also benefits from this line of research. \cite{wei2018task} applies DQN to decide whether to collect more symptom information from patients by continuing the conversation or conclude the diagnosis with predicting a disease. \cite{xu2019end} proposes a knowledge-routed DQN to integrate medical knowledge into dialog management. By considering the relations among diseases and symptoms during decision making, the accuracy of disease prediction is improved. In this work, we integrate meta-learning with DQN to improve the learning efficiency, and evaluate the proposed dialog manager on medical domain.

\subsection{Meta-Learning}
% Despite the progress in deep RL for goal oriented dialog systems, it is prohibitively expensive and time-consuming to build such systems for the requirement of a large number of labeled examples to train RL models, especially in the tasks require domain-specific knowledge, for example, healthcare. Another challenge to both human doctor and computer is learning and recognizing new diseases, when we do not have enough data about it to learn from.

Recently meta learning starts gaining attention among the whole machine learning field for improving model performance when few labeled training data is available. Model-Agnostic Meta-Learning \cite{finn2017model} optimizes parameter initialization over multiple out-of-domain subtasks for the initialization to be generalizable in targeted tasks after fine-tuning on few in-domain labels. Neural Turing machines \cite{graves2014neural} augment neural models with memory modules to improve performance in limited-data regime. Metric-based meta learning, such as prototypical networks (ProtoNets) \cite{snell2017prototypical}, siamese neural networks \cite{koch2015siamese}, and matching networks \cite{vinyals2016matching}, learns embedding or metric space such that the space can be adapted to domains unseen in the training set with few examples from the unseen domains. Meta-learning models have also been applied in dialog generation \cite{qian2019domain} for quick policy adaption in different dialog domains. In this work, our proposed model is evaluated on the medical domain that requires not only dialog policy, but also multi-step reasoning with domain-specific knowledge.

\section{Method}

\subsection{Dialog State Representations}
\label{sec:dsr}
Following the method proposed in \cite{wei2018task} for vectorizing the dialog states, each dialog state consists of 4 parts:

\noindent \textbf{I. UserAction}: The user action of the previous dialog turn:
\begin{itemize}[noitemsep,topsep=0pt]
\item \textbf{Request}: A user sends a self-report containing a set of explicit symptoms and request for diagnosis.
\item \textbf{Confirm}: A user confirms the existence of an agent-inquired symptom.
\item \textbf{Deny}: A user denies the existence of a symptom.
\item \textbf{NotSure}: A user is not sure about the inquired symptom, usually happened when an unrelated symptom is inquired.
\end{itemize}
\textbf{II. AgentAction}: The previous action of the dialog agent:
\begin{itemize}[noitemsep,topsep=0pt]
\item \textbf{Initiate}: The agent initiates the dialog and asks the user to send the self-report.
\item \textbf{Request}: The agent asks the user if a symptom exists.
\item \textbf{Inform}: The system predicts and inform the disease.
\end{itemize}
\textbf{III. Slots}: The set of symptoms appeared in the dialog history and their status. Each symptom has 4 possible status,
\begin{itemize}[noitemsep,topsep=0pt]
\item \textbf{Confirmed}: Existence of the symptom is confirmed.
\item \textbf{Denied}: Existence is denied by the user.
\item \textbf{Unrelated}: The symptom is not necessary for the doctor to make an accurate diagnosis.
\item \textbf{NotInquired}: The symptom has not been inquired.
\end{itemize}
\textbf{IV. NumTurns}: The length of the dialog history, in other words, current number of turns.

In each dialog turn, we represent UserAction, AgentAction, and NumTurns with one-hot vectors $a^u, a^r$, and $n$ respectively. We use a 66-dimension vector $s$ to represent the Slots, where each dimension indicates the status of a symptom. A confirmed, denied, unrelated, and not inquired symptom possesses value $1$, $-1$, $-2$, and $0$ in the corresponding dimension. The final input of the neural networks at the $t$-th turn is represented as
\begin{equation}
\label{eq:state}
s_t = [a_t^u, a_t^r, n_t, s_t]
\end{equation}

\subsection{Prototypical Q Networks}
In this work, we propose the prototypical Q networks (ProtoQNs) as well as corresponding training and evaluation pipelines, based on conventional DQNs and ProtoNets.

We first define the notation of the our dataset as follows. The dataset $S$ contains $N$ doctor-patient conversations, $\left \{ (x_1, y_1), (x_2, y_2), \dots, (x_N, y_N) \right \}$, where $y_i$ stands for the disease label of the $i-$th training case and $x_i$ stands for the corresponding dialog history, and $x_i$ can be further represented as
\begin{equation}
\label{eq:te}
x_i = \left \{u_0^i = E^i, (a_1^i, u_1^i), \dots, (a_k^i, u_k^i), a_{k+1}^i = D^i \right \}
\end{equation}
Here $E$ stands for explicit symptoms reported at the beginning of each dialog, $a$ stands for agent follow-up inquiries, $u$ stands for user responses, and $D$ stands for the predicted disease.

The core of the ProtoNets \cite{snell2017prototypical} is calculating and updating the prototype embeddings of the output classes. The network classifies examples by comparing the input embedding with the prototypes, and predict the class with its prototype closest to the input embedding. As compared to other single-stage reasoning task, such as image classification or object detection \cite{snell2017prototypical}, medical dialog requires multi-stage reasoning to infer dialog action through states. As a result, we propose a method for computing action prototype via dialog state embedding.

\subsubsection{Dialog state embedding}
% The main feature of the prototypical networks \cite{snell2017prototypical} is calculating and updating the prototype embeddings of the output classes. The network conduction the classification by comparing the input embedding and prototypical embeddings, and predict the class with closest embedding to the input enbedding. In this work, we propose a method for calculating prototyppe embeddings for dialog actions.

% Different from other single-step reasoning task, for example image classification or object detection \cite{snell2017prototypical}, medical dialog is a multi-step reasoning process. As a result, we encode dialog actions with dialog states. For each real conversation in the training set $x_i$ defined in Equation \ref{eq:te}, we represent each dialog state $s_k$, including symptom queries $a_j^i$ and disease predictions with previous dialog state by

For each training conversation, $x_i$, as defined in Equation \ref{eq:te}, the dialog state $s_j^i$ at time step $j$ can be represented as
\begin{equation}
s_j^i = \left \{ u_0, (a_1^i, u_1^i), \dots, (a_{j-1}^i, u_{j-1}^i) \right \}
\end{equation}
$s_j^i$ is then converted into a representation vector following Equation \ref{eq:state}, denoted as $f_{enc}$, to obtain the state embedding, $e_j^i$. That is
\begin{equation}
\label{eq:se}
e_j^i = f_{enc} (s_j^i)
\end{equation}
With the approach described above, for any conversation we can get $e_t$, the embedding of the dialog state at time step t (i.e., $s_t$).
\subsubsection{Dialog action prediction}
At each step of a dialog, the dialog system is provided a dialog state encoding $s_t$ in Equation \ref{eq:state}. With the same encoder $f_{enc}$ for embedding dialog states in the training set, we calculate the embedding of the input dialog state $e_t$ in the new dialogs generated in both training and evaluation phases with Equation \ref{eq:se}.
%
% \begin{equation}
% e_t = f_{enc} (s_t)
% \end{equation}
%

Then we can further compute prototypes and predict dialog action. First, with the state embedding $e_t$, the protoQNs calculate the Q value of the $m$-th dialog actions $a_m$ by
\begin{equation}
\label{eq:q}
q(a = a_m) = e_t \cdot P_m
\end{equation}
where $a_m$ is the embedding of the $m$-th dialog action, generated by mean-pooling a number of dialog states followed by $a_m$, and
\begin{equation}
\label{eq:proto}
P_m = \frac{ \sum_{i, j \in D} v(a_j^i) \cdot \mathds{1} (a_j^i = a_m) } { \sum_{i, j \in D}  \mathds{1} (a_j^i = a_m) }
\end{equation}
Here, $\mathds{1} (\cdot)$ is the indicator function, and $D$ is the set of examples used for computing prototypes. In training, $D$ is a small subset of dialog states sampled from training set follwed by action $a_m$, while in evaluation, $D$ is the entire training set. In Algorithm \ref{alg:protoqn}, we show the calculation of prototype embeddings.
\begin{algorithm}
\caption{Calculating prototype embeddings}
\textbf{Function} $protoEmbed(A, H)$\\
\textbf{Inputs:} dialog action set $A$, entire dialog history in the training corpus $H = \left \{(s_1, a_1), \dots, (s_m, a_m)\right \}$.\\
\textbf{Outputs:} Prototype embeddings of dialog actions $P$.
\begin{algorithmic}[1]
\For {$a' \in A$}
\State $A_{a'} = \left \{(s_k, a_k) \in H, a_k = a'\right \}$
\If {Training}
\State Sample D from $A_{a'}$ with size = 10
\ElsIf {Evaluating}
\State D = $A_{a'}$
\EndIf
\State Calculate $P_{a'}$ with Equation \ref{eq:proto}
\EndFor
\end{algorithmic}
\label{alg:protoqn}
\end{algorithm}

\subsubsection{Disease prediction and training}
Each dialog starts from an explicit symptom set provided by a user goal, and the model inquiries a set of symptoms before making the final disease prediction. For each inquiry, the user simulator replies based on the implicit symptom set as described in Section \ref{sec:dsr}. The conversation stops when the systems output a disease prediction. Summarizing the previous sections and descriptions, we provide the complete procedure of a medical dialog in Algorithm \ref{alg:dialog}.

For each simulated dialog described above, the model sees a success or failure reward when it informs the user a predicted disease. The ProtoQN updates its weights based on the reward with stochastic gradient descent (SGD) following the standard pipeline of training a DQN applied in \cite{wei2018task,xu2019end}. For evaluation, the ProtoQN generates prototype embeddings only once before the evaluation begins with all real dialog histories in the training corpus.
%The model initicate the dialog by given an explicit symptom set from the user, and inquiries for a set of symptoms before making the final disease prediction. Summarizing the previous sections and descriptions, we describe the complete procedure of a medical dialog in Algorithm \ref{alg:dialog}.
%
\begin{algorithm}
\caption{Automatic diagnosis dialog process}
\textbf{Inputs:} Explicit symptom set $E$, initial turn Id $t=1$, empty implicit symptom set $I = \left \{\right \}$, dialog history for training $H$, encoder $f_{enc}$\\
\textbf{Outputs:} Final disease prediction $D$.
\begin{algorithmic}[1]
\State $P = protoEmbed(A, H)$
\While {Dialog not end}
\State $s_t = State(E, I, t)$ with Equation \ref{eq:state}
\State $h_t = f_{enc} (s_t)$
\State $a_t = \argmax_a h_t \cdot P_a^T$ with Equation \ref{eq:q}
\If {$a \in$ Symptom}
\State $u_t = $ user response
\State $I = I + (a_t, u_t)$
\State $t = t + 1$
\ElsIf {$a \in$ Disease}
\State $D = a_t$
\State $endDialog()$
\EndIf
\EndWhile
\end{algorithmic}
\label{alg:dialog}
\end{algorithm}
\begin{figure*}[t!]
    \centering
    \begin{subfigure}[t]{0.24\textwidth}
        \centering
        \includegraphics[height=1in]{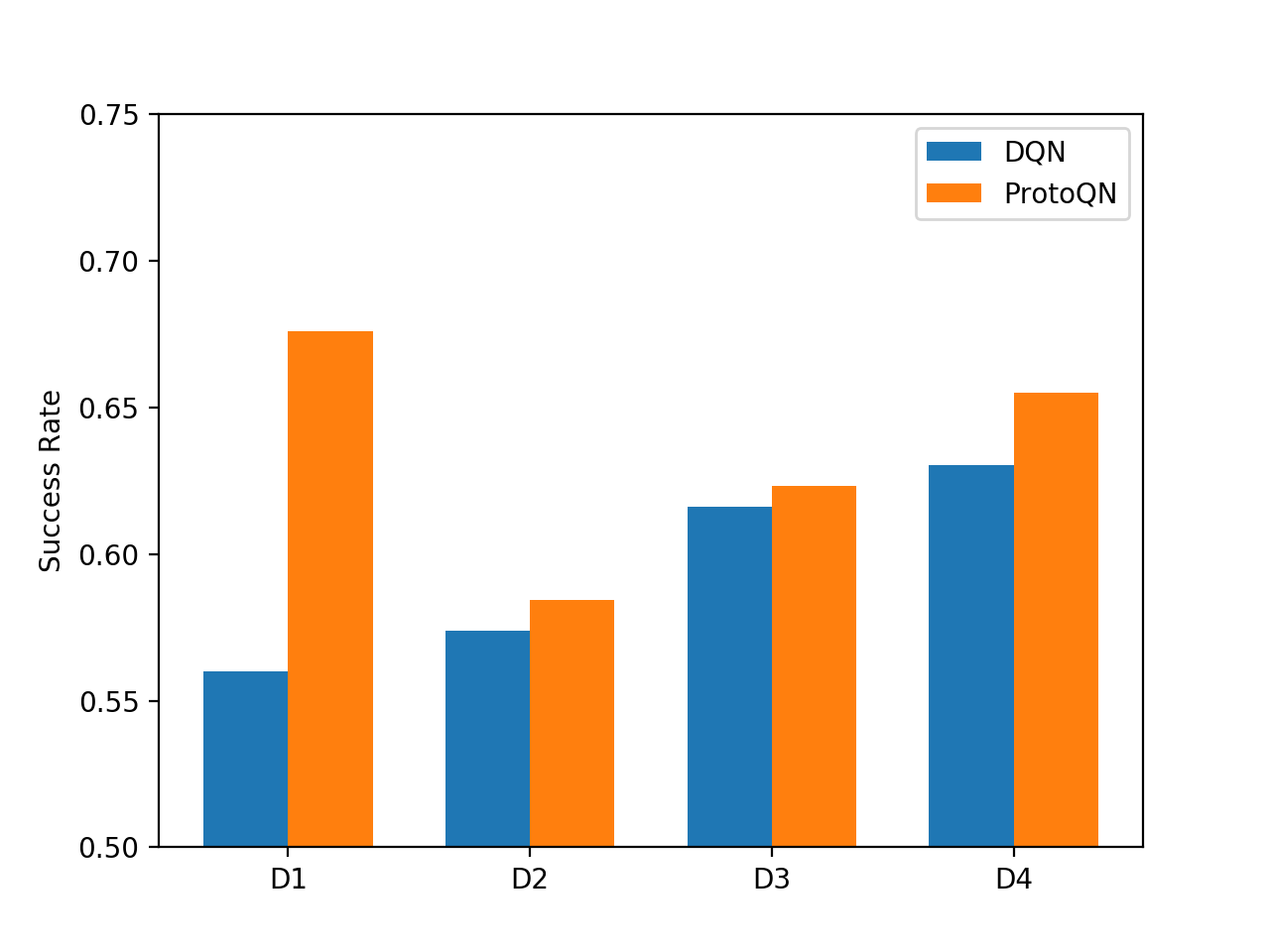}
        \caption{Noise $= 0.0$}
        \label{fig:step:1}
    \end{subfigure}
    ~ 
    \begin{subfigure}[t]{0.24\textwidth}
        \centering
        \includegraphics[height=1in]{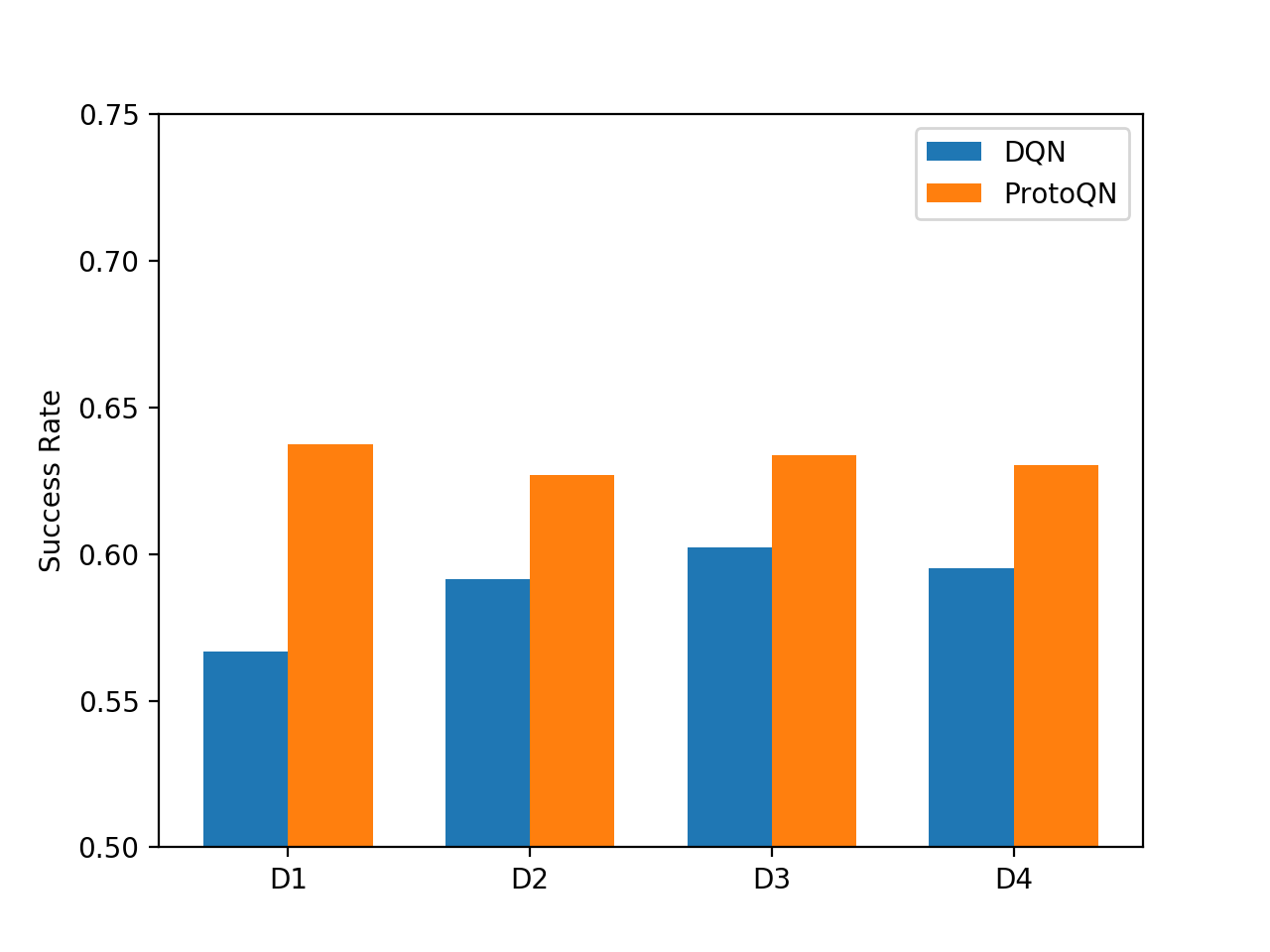}
        \caption{Noise $= 0.1$}
        \label{fig:step:2}
    \end{subfigure}
    ~ 
    \begin{subfigure}[t]{0.24\textwidth}
        \centering
        \includegraphics[height=1in]{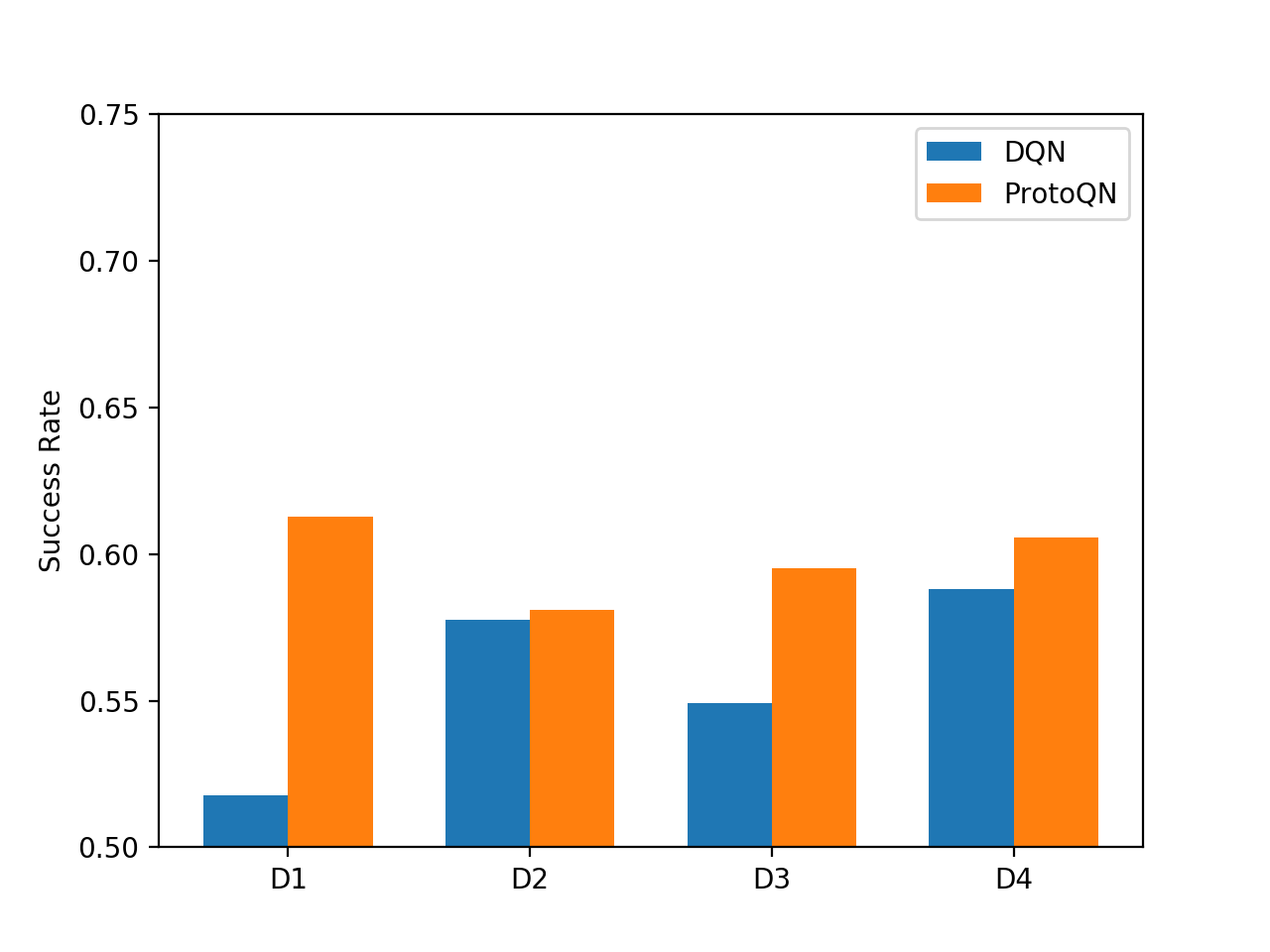}
        \caption{Noise $= 0.2$}
        \label{fig:step:3}
    \end{subfigure}
    ~
    \begin{subfigure}[t]{0.24\textwidth}
        \centering
        \includegraphics[height=1in]{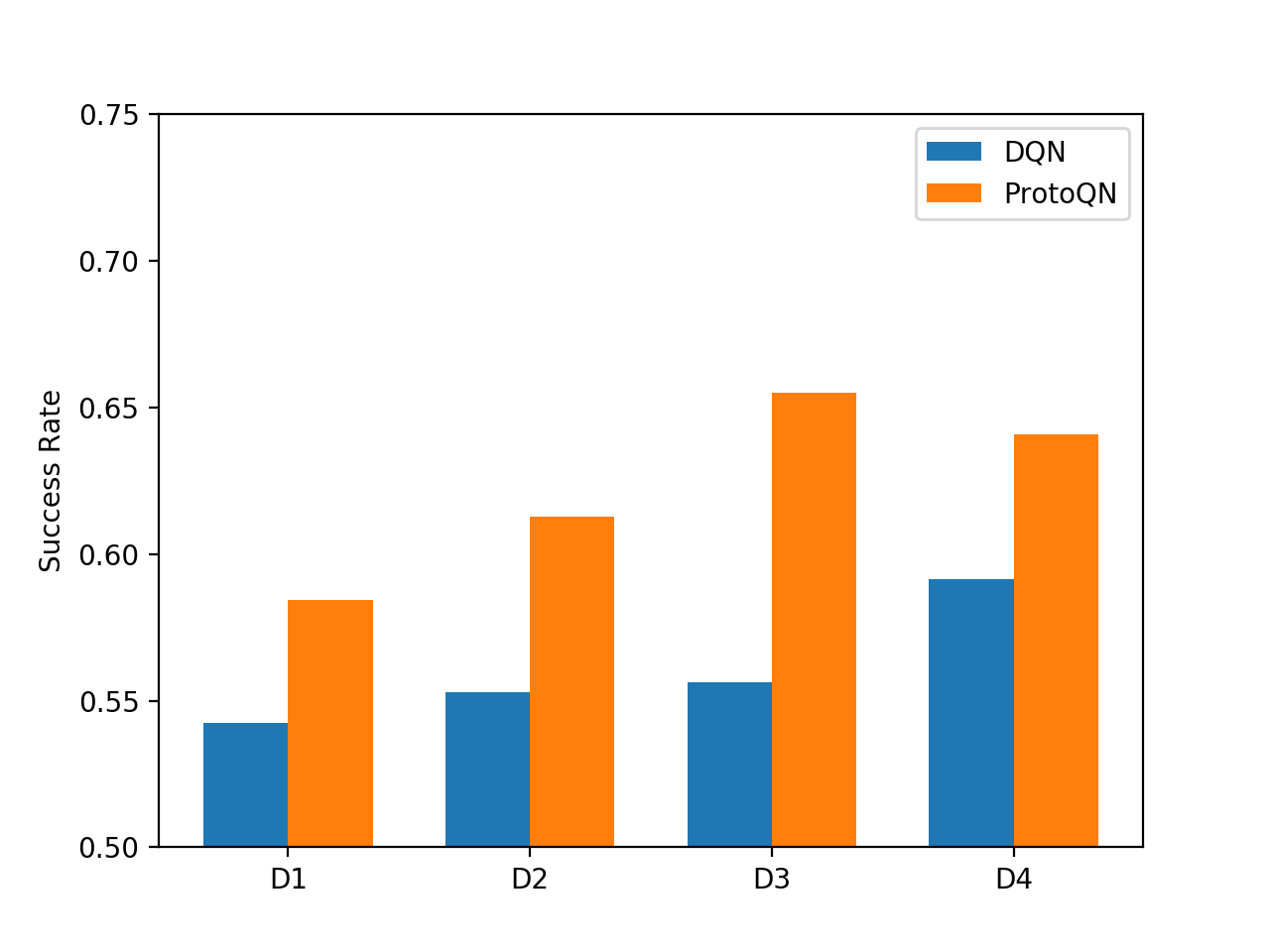}
        \caption{Noise $= 0.3$}
        \label{fig:step:4}
    \end{subfigure}
    
    \caption{Performances of ProtoQN and DQN on 4 noise levels. Each figure stands for a noise, while the blue bars stand for the performances of DQNs, and the orange bars stand for the performances of ProtoQNs.}
    \label{fig:gmemnn}
\end{figure*}

\section{Experiments}
\subsection{Data and Experiment Settings}
In this work, we evaluate the models on Muzhi dataset \cite{wei2018task}. The dataset contains 710 medical dialogs between real doctors and patients, and each is annotated as a user goal, covering 4 different diseases and 66 symptoms. In this work, we apply the official train-test split. 568 user goals are used from training and other 142 are used for evaluation.

Meanwhile, we apply a simulator for providing user response in the conversations. To simulate speech noises and mistakes led by user knowledge biases, we apply intent and slot noises to the simulator \cite{xu2019end} to evaluate the model performances under different levels of noises. In our experiments, we apply 0\%, 10\%, 20\%, and 30\% error rates, i.e., the probability that the status of an inquired symptom is sampled at random rather than based on annotation.

We also conduct 2 groups of experiments for evaluating the model performances at various conditions. Firstly we train both ProtoQN and DQN on the entire training set. This is the conventional with fully supervised learning setting. Secondly, we adopt a meta-learning-like setting to evaluate the model performance at few-shot learning. We pre-train the models with 3 diseases, and fine-tune the models with randomly selected training samples from the trained diseases plus a few cases of the new disease. In both learning tasks, the neural models are evaluated on the entire public test set. In this paper we set the success reward of ProtoQN to 20, failure reward to 0, and the maximum number of turns is 44. For DQN, we keep the settings in \cite{wei2018task}.

\subsection{Fully Supervised Learning}
We first compare the ProtoQN with DQN at the normal supervised learning setting to evaluate their abilities of learning dialog policy with enough training data. Experimental results are shown in Table \ref{tab:exp1}. We obtain the DQN performance directly from \cite{wei2018task}.
\begin{table}[h]
\centering
\caption{Experimental results of ProtoQN and DQN on supervised learning.}
\begin{tabular}{@{}lll@{}}
\toprule
Models  & Success Rate (\%) & Reward \\ \midrule
DQN     & 65         & 20.51  \\
ProtoQN & 70.42       & 23.58  \\ \bottomrule
\end{tabular}
\label{tab:exp1}
\end{table}

Experiments showed that the ProtoQN significantly outperformed the DQN baseline, without adding external knowledge and hand-crafted features \cite{xu2019end}. The improvement showed that utilizing the dialog history between real doctors and patients to calculate prototype embeddings is effective and provides better ability for the model to learn the dialog policy. While DQN learns Q-values indirectly from simulated conversations, ProtoQN directly relates Q-values with dialog actions and states in real conversation.

\subsection{Few-shot New Disease Adaption}
In this part, we investigate the model performance at the situation when a new disease appears after the model was trained, and only a few examples are available for training. To evaluate this situation, we adopt an experiment setting popular in few-shot learning, where we first pre-train the models on the training set from a subset of diseases. Then the models are fine-tuned (i.e., adapted) on a small number, $N$, of examples from a new disease. In order to prevent catastrophic forgetting \cite{riemer2018learning}, we also randomly select $N$ samples from each pre-trained diseases to compose the adaptation set. Since the Muzhi corpus includes 4 diseases, we conduct 4 iterations of evaluation, with each corresponding to one of the 4 diseases as the new disease, and the remaining as the pre-trained ones. After fine-tuning, the models are also evaluated with the public test set. Also, we consider different noise levels in this task. Our purpose is evaluating how well the models learn new diseases with few examples and not forgetting the knowledge for the pre-trained diseases.
\begin{table}[]
\centering
\caption{Average success rates (\%) of the meta-learning tasks with DQN and ProtoQN under different noise levels.}
\begin{tabular}{@{}lllll@{}}
\toprule
Noise   & 0     & 0.1   & 0.2   & 0.3   \\ \midrule
DQN     & 59.51 & 58.89 & 55.81 & 56.07 \\
ProtoQN & 63.47 & 63.21 & 59.85 & 62.32 \\ \bottomrule
\end{tabular}
\label{tab:meta}
\end{table}
%
% \begin{table}[]
% \begin{tabular}{@{}lllllllllll@{}}
% \toprule
% Disease & \multicolumn{2}{c}{1} & \multicolumn{2}{c}{2} & \multicolumn{2}{c}{3} & % \multicolumn{2}{c}{4} & \multicolumn{2}{c}{Average} \\ \midrule
% Noise   & DQN      & ProtoQN    & DQN      & ProtoQN    & DQN      & ProtoQN    & DQN      & ProtoQN    & DQN         & ProtoQN       \\
% 0       & 55.99    & 67.61      & 57.39    & 58.45      & 61.62    & 60.92      & 63.03    & 65.50      & 59.51       & 63.12         \\
% 0.1     & 56.69    & 63.73      & 59.12    & 62.68      & 60.22    & 63.38      & 59.51    & 63.03      & 58.89       & 63.21         \\
% 0.2     & 51.76    & 61.27      & 57.75    & 57.04      & 54.93    & 59.51      & 58.80    & 60.56      & 55.81       & 59.59         \\
% 0.3     & 54.23    & 58.45      & 55.28    & 61.27      & 55.64    & 65.49      & 59.16    & 64.09      & 56.07       & 62.32         \\ \bottomrule
% \end{tabular}
% \caption{}
% \label{tab:meta}
% \end{table}

The experiment results of the new disease adaption are shown in Table \ref{tab:meta}. From the table we see that the ProtoQN significantly outperformed the DQN model and achieves SOTA performance under few-shot learning setting. The fact shows that learning shared embeddings from real conversations is more efficient when adapting the model for learning new diseases with few examples. Meanwhile, many shared knowledge from pre-trained diseases can still be applied for the new ones. The learned knowledge about symptoms allows the neural network to learn new diseases with a better initialization of dialog policy, and thus makes the adaptation faster and better. We also found that although the increasing of noise level degrades the performance of DQNs, ProtoQN yields steady success rate when the noise level varies. we believe the robustness results from the ensemble nature of ProtoNet’s inference mechanisms.

\section{Conclusion and Future Work}
In this work, we propose a novel dialog management model, prototypical Q network, for supervised and few-shot dialog policy learning. We apply this model in the area of automatic conversational diagnosis. Experiments showed that the ProtoQNs outperforms the DQN model in both supervised and few-shot settings. In the supervised setting, ProtoQNs achieve results comparable to SOTA without using domain-specific features. As for the few-shot experiment, ProtoQN learns new diseases using few training samples without forgetting previously learned, and achieves SOTA. The model also shows less degradation as we injecting noise to conversation. Our study suggests that modeling real conversations directly reinforces simulator-based dialog policy learning. Embeddings of dialog actions are shareable among tasks (diseases, in our case) and benefits the fast adaptation to new ones. Here we show promising results in medical domain. In future, we will investigate more adaptive models as well as different domains and corpora toward the goal of modeling new dialog tasks better and with fewer examples.

\bibliographystyle{IEEEtran}

\bibliography{mybib}

\begin{thebibliography}{10}
\providecommand{\url}[1]{#1}
\csname url@samestyle\endcsname
\providecommand{\newblock}{\relax}
\providecommand{\bibinfo}[2]{#2}
\providecommand{\BIBentrySTDinterwordspacing}{\spaceskip=0pt\relax}
\providecommand{\BIBentryALTinterwordstretchfactor}{4}
\providecommand{\BIBentryALTinterwordspacing}{\spaceskip=\fontdimen2\font plus
\BIBentryALTinterwordstretchfactor\fontdimen3\font minus
  \fontdimen4\font\relax}
\providecommand{\BIBforeignlanguage}[2]{{%
\expandafter\ifx\csname l@#1\endcsname\relax
\typeout{** WARNING: IEEEtran.bst: No hyphenation pattern has been}%
\typeout{** loaded for the language `#1'. Using the pattern for}%
\typeout{** the default language instead.}%
\else
\language=\csname l@#1\endcsname
\fi
#2}}
\providecommand{\BIBdecl}{\relax}
\BIBdecl

\bibitem{wei2018task}
Z.~Wei, Q.~Liu, B.~Peng, H.~Tou, T.~Chen, X.~Huang, K.-F. Wong, and X.~Dai,
  ``Task-oriented dialogue system for automatic diagnosis,'' in
  \emph{Proceedings of the 56th Annual Meeting of the Association for
  Computational Linguistics (Volume 2: Short Papers)}, 2018, pp. 201--207.

\bibitem{mnih2013playing}
V.~Mnih, K.~Kavukcuoglu, D.~Silver, A.~Graves, I.~Antonoglou, D.~Wierstra, and
  M.~Riedmiller, ``Playing atari with deep reinforcement learning,''
  \emph{arXiv preprint arXiv:1312.5602}, 2013.

\bibitem{xu2019end}
L.~Xu, Q.~Zhou, K.~Gong, X.~Liang, J.~Tang, and L.~Lin, ``End-to-end
  knowledge-routed relational dialogue system for automatic diagnosis,''
  \emph{arXiv preprint arXiv:1901.10623}, 2019.

\bibitem{finn2017model}
C.~Finn, P.~Abbeel, and S.~Levine, ``Model-agnostic meta-learning for fast
  adaptation of deep networks,'' in \emph{Proceedings of the 34th International
  Conference on Machine Learning-Volume 70}.\hskip 1em plus 0.5em minus
  0.4em\relax JMLR. org, 2017, pp. 1126--1135.

\bibitem{snell2017prototypical}
J.~Snell, K.~Swersky, and R.~Zemel, ``Prototypical networks for few-shot
  learning,'' in \emph{Advances in neural information processing systems},
  2017, pp. 4077--4087.

\bibitem{vinyals2016matching}
O.~Vinyals, C.~Blundell, T.~Lillicrap, D.~Wierstra \emph{et~al.}, ``Matching
  networks for one shot learning,'' in \emph{Advances in neural information
  processing systems}, 2016, pp. 3630--3638.

\bibitem{silver2016mastering}
D.~Silver, A.~Huang, C.~J. Maddison, A.~Guez, L.~Sifre, G.~Van Den~Driessche,
  J.~Schrittwieser, I.~Antonoglou, V.~Panneershelvam, M.~Lanctot \emph{et~al.},
  ``Mastering the game of go with deep neural networks and tree search,''
  \emph{nature}, vol. 529, no. 7587, p. 484, 2016.

\bibitem{zhao2016towards}
T.~Zhao and M.~Eskenazi, ``Towards end-to-end learning for dialog state
  tracking and management using deep reinforcement learning,'' \emph{arXiv
  preprint arXiv:1606.02560}, 2016.

\bibitem{yang2017end}
X.~Yang, Y.-N. Chen, D.~Hakkani-T{\"u}r, P.~Crook, X.~Li, J.~Gao, and L.~Deng,
  ``End-to-end joint learning of natural language understanding and dialogue
  manager,'' in \emph{2017 IEEE International Conference on Acoustics, Speech
  and Signal Processing (ICASSP)}.\hskip 1em plus 0.5em minus 0.4em\relax IEEE,
  2017, pp. 5690--5694.

\bibitem{lipton2016efficient}
Z.~C. Lipton, J.~Gao, L.~Li, X.~Li, F.~Ahmed, and L.~Deng, ``Efficient
  exploration for dialogue policy learning with bbq networks \& replay buffer
  spiking,'' \emph{arXiv preprint arXiv:1608.05081}, vol.~3, 2016.

\bibitem{peng2017composite}
B.~Peng, X.~Li, L.~Li, J.~Gao, A.~Celikyilmaz, S.~Lee, and K.-F. Wong,
  ``Composite task-completion dialogue policy learning via hierarchical deep
  reinforcement learning,'' \emph{arXiv preprint arXiv:1704.03084}, 2017.

\bibitem{lin1992self}
L.-J. Lin, ``Self-improving reactive agents based on reinforcement learning,
  planning and teaching,'' \emph{Machine learning}, vol.~8, no. 3-4, pp.
  293--321, 1992.

\bibitem{papineni2001natural}
K.~A. Papineni, S.~Roukos, and R.~T. Ward, ``Natural language task-oriented
  dialog manager and method,'' Jun.~12 2001, uS Patent 6,246,981.

\bibitem{scheffler2002automatic}
K.~Scheffler and S.~Young, ``Automatic learning of dialogue strategy using
  dialogue simulation and reinforcement learning,'' in \emph{Proceedings of the
  second international conference on Human Language Technology Research}.\hskip
  1em plus 0.5em minus 0.4em\relax Morgan Kaufmann Publishers Inc., 2002, pp.
  12--19.

\bibitem{young2010hidden}
S.~Young, M.~Ga{\v{s}}i{\'c}, S.~Keizer, F.~Mairesse, J.~Schatzmann,
  B.~Thomson, and K.~Yu, ``The hidden information state model: A practical
  framework for pomdp-based spoken dialogue management,'' \emph{Computer Speech
  \& Language}, vol.~24, no.~2, pp. 150--174, 2010.

\bibitem{young2013pomdp}
S.~Young, M.~Ga{\v{s}}i{\'c}, B.~Thomson, and J.~D. Williams, ``Pomdp-based
  statistical spoken dialog systems: A review,'' \emph{Proceedings of the
  IEEE}, vol. 101, no.~5, pp. 1160--1179, 2013.

\bibitem{silver2014deterministic}
D.~Silver, G.~Lever, N.~Heess, T.~Degris, D.~Wierstra, and M.~Riedmiller,
  ``Deterministic policy gradient algorithms,'' 2014.

\bibitem{seneff2000dialogue}
S.~Seneff and J.~Polifroni, ``Dialogue management in the mercury flight
  reservation system,'' in \emph{Proceedings of the 2000 ANLP/NAACL Workshop on
  Conversational systems-Volume 3}.\hskip 1em plus 0.5em minus 0.4em\relax
  Association for Computational Linguistics, 2000, pp. 11--16.

\bibitem{dodge2015evaluating}
J.~Dodge, A.~Gane, X.~Zhang, A.~Bordes, S.~Chopra, A.~Miller, A.~Szlam, and
  J.~Weston, ``Evaluating prerequisite qualities for learning end-to-end dialog
  systems,'' \emph{arXiv preprint arXiv:1511.06931}, 2015.

\bibitem{bordes2016learning}
A.~Bordes, Y.-L. Boureau, and J.~Weston, ``Learning end-to-end goal-oriented
  dialog,'' \emph{arXiv preprint arXiv:1605.07683}, 2016.

\bibitem{chattopadhyay2017evaluating}
P.~Chattopadhyay, D.~Yadav, V.~Prabhu, A.~Chandrasekaran, A.~Das, S.~Lee,
  D.~Batra, and D.~Parikh, ``Evaluating visual conversational agents via
  cooperative human-ai games,'' in \emph{Fifth AAAI Conference on Human
  Computation and Crowdsourcing}, 2017.

\bibitem{graves2014neural}
A.~Graves, G.~Wayne, and I.~Danihelka, ``Neural turing machines,'' \emph{arXiv
  preprint arXiv:1410.5401}, 2014.

\bibitem{koch2015siamese}
G.~Koch, R.~Zemel, and R.~Salakhutdinov, ``Siamese neural networks for one-shot
  image recognition,'' in \emph{ICML deep learning workshop}, vol.~2.\hskip 1em
  plus 0.5em minus 0.4em\relax Lille, 2015.

\bibitem{qian2019domain}
K.~Qian and Z.~Yu, ``Domain adaptive dialog generation via meta learning,''
  \emph{arXiv preprint arXiv:1906.03520}, 2019.

\bibitem{riemer2018learning}
M.~Riemer, I.~Cases, R.~Ajemian, M.~Liu, I.~Rish, Y.~Tu, and G.~Tesauro,
  ``Learning to learn without forgetting by maximizing transfer and minimizing
  interference,'' \emph{arXiv preprint arXiv:1810.11910}, 2018.

\end{thebibliography}

% \begin{thebibliography}{9}
% \bibitem[1]{Davis80-COP}
%   S.\ B.\ Davis and P.\ Mermelstein,
%   ``Comparison of parametric representation for monosyllabic word recognition in continuously spoken sentences,''
%   \textit{IEEE Transactions on Acoustics, Speech and Signal Processing}, vol.~28, no.~4, pp.~357--366, 1980.
% \bibitem[2]{Rabiner89-ATO}
%   L.\ R.\ Rabiner,
%   ``A tutorial on hidden Markov models and selected applications in speech recognition,''
%   \textit{Proceedings of the IEEE}, vol.~77, no.~2, pp.~257-286, 1989.
% \bibitem[3]{Hastie09-TEO}
%   T.\ Hastie, R.\ Tibshirani, and J.\ Friedman,
%   \textit{The Elements of Statistical Learning -- Data Mining, Inference, and Prediction}.
%   New York: Springer, 2009.
% \bibitem[4]{YourName17-XXX}
%   F.\ Lastname1, F.\ Lastname2, and F.\ Lastname3,
%   ``Title of your INTERSPEECH 2020 publication,''
%   in \textit{Interspeech 2020 -- 20\textsuperscript{th} Annual Conference of the International Speech Communication Association, September 15-19, Graz, Austria, Proceedings, Proceedings}, 2020, pp.~100--104.
% \end{thebibliography}

\end{document}